\documentclass[11pt,a4paper]{article}
\usepackage{coling2020}
\usepackage{times}
\usepackage{latexsym}
\usepackage{booktabs}
\usepackage{url}
\usepackage{tabularx}
\usepackage{amsmath}
\usepackage{graphicx}
\usepackage{dblfloatfix} 
\usepackage{url}
\usepackage{hyperref}
\usepackage{booktabs}
\usepackage{tablefootnote}
\usepackage{caption}
\colingfinalcopy

\title{SemEval-2020 Task 6: Definition extraction from free text with the DEFT corpus}

\author{Sasha Spala$^1$, Nicholas A Miller$^1$, Franck Dernoncourt$^2$, Carl Dockhorn$^1$ \\
  $^1$Adobe Inc., $^2$Adobe Research \\
  345 Park Ave\\
  San Jose, CA 95110-2704 \\
  {\tt \{sspala,nimiller,dernonco,cdockhorn\}@adobe.com}  \\
}
\date{}

\begin{document}
\maketitle
\begin{abstract}
  Research on definition extraction has been conducted for well over a decade, largely with significant constraints on the type of definitions considered. In this work, we present DeftEval, a SemEval shared task in which participants must extract definitions from free text using a term-definition pair corpus that reflects the complex reality of definitions in natural language. Definitions and glosses in free text often appear without explicit indicators, across sentences boundaries, or in an otherwise complex linguistic manner. DeftEval involved 3 distinct subtasks: 1) Sentence classification, 2) sequence labeling, and 3) relation extraction.
\end{abstract}

\section{Introduction}

Definition extraction as a complex, real-world task is currently an emerging field of study. Traditional definition extraction approaches mostly rely on simple, syntactically straight-forward examples with relatively little variance in vocabulary. Corpora, including the WCL \cite{Navigli_data:2010} and ukWaC \cite{Ferraresi:2008}, typically consist of ``definition sentences" which follow a standard \textit{X is a (type) Y} or \textit{X, such as Y} syntactic structure. Many also contain definitors \cite{Navigli:2010}, such as ``means", ``is", or ``is defined by". We provide a complete review of the existing state of definition extraction in \newcite{spala:19}.

\section{Data}

\begin{table*}[t]
\begin{tabular}{p{3.5cm}p{11.5cm}}
\toprule
Tag Name               & Description \\ \midrule
Term                   & A primary term           \\
Alias Term             & A secondary, less common name for the primary term. Links to a term tag.         \\
Ordered Term           & Multiple terms that have matching sets of definitions which cannot be separated from each other without creating an non-contiguous sequence of tokens. E.g. \textit{x and y represent positive and negative versions of the definition, respectively}           \\
Referential Term       & An NP reference to a previously mentioned term tag. Typically \textit{this/that/these + NP}          \\
Definition             & A primary definition of a term. May not exist without a matching term.           \\
Secondary Definition   & Supplemental information that may qualify as a definition sentence or phrase, but crosses a sentence boundary.           \\
Ordered Definition     & Multiple definitions that have matching sets of terms which cannot be separated from each other. See Ordered Term.           \\
Referential Definition & NP reference to a previously mentioned definition tag. See Referential Term. \\
Qualifier & A specific date, location, or condition under which the definition holds \\
\bottomrule
\end{tabular}
\caption{Tag schema}
\label{tag_schema}
\end{table*}
\begin{table*}[t]
\begin{tabular}{p{3.5cm}p{11.5cm}}
\toprule
Relation Name               & Description \\ \midrule
Direct-defines                   & Links definition to term.           \\
Indirect-defines             & Links definition to referential term or term to referential definition.      \\
Refers-to & Links referential term to term or referential definition to definition. \\
AKA           & Links alias term to term.           \\
Supplements      & Links secondary definition to definition.           \\ \bottomrule
\end{tabular}
\caption{Relation schema}
\label{relation_schema}
\end{table*}

The DeftEval task provided data from the DEFT corpus for training, development and testing. Introduced in Spala et al \shortcite{spala:19}, the DEFT corpus is currently the largest and most comprehensive corpus explicitly for definition extraction. In addition to the typical definition-type sentences discussed in the introduction, the corpus also contains sentence and ``sentence windows" of wide variance in syntactic and semantic construction. Sentences for the corpus were retrieved from the open source textbook website \href{https://cnx.org}{cnx.org}, as well as from various 2017 SEC contract filings from the publicly available \href{https://www.sec.gov/cgi-bin/srch-edgar}{US Securities and Exchange Commission EDGAR (SEC) database}. For the DeftEval shared task, participants were only required to use sentences extracted from textbooks. EDGAR annotations were provided on request, but not required for participation in the shared task.

The DEFT corpus provides sentence ``context windows" for textbook data from \href{https://cnx.org}{cnx.org} to narrow the search space for possible definitions. A context window is defined as a set of three sentences: one sentence before a sentence containing a potential term (e.g. a bold word), the main sentence with the potential term, and one sentence after. Though it is certainly possible that some definitions may occur further away from the bolded mention of the term, for the purposes of the shared task, definitions that occur outside of the three-sentence window around a potential term are out of scope.

The corpus contains 9 different token labels (see Table\ref{tag_schema}), and 5 different types of relations (see Table\ref{relation_schema}). This expanded annotation schema allows for a wide range of data collection, as well as a better understanding of ``non-traditional" definition types (i.e. beyond the typical \textit{x is a y} structure). With this annotation, it is possible to observe long-distance (e.g. cross-sentence) term-definition relationships, as well as supplemental definition content.
\textit{Term} and \textit{definition} tags in the schema are treated largely similar to Storrer and Wellinghoff's \textit{definiens} and \textit{definitor} tags, respectively \shortcite{Storrer:2006}. A term is defined as a primary term in a sentence which is accompanied by a definition phrase or sentence within the sentence context window. Terms are most commonly expressed as noun phrases:
\begin{quote}
    \textit{The scientific method} is a method of research with defined steps that include experiments and careful observation.\footnote{\href{https://github.com/adobe-research/deft_corpus/blob/master/data/deft_files/dev/t1_biology_0_0.deft}{https://github.com/adobe-research/deft\_corpus/blob/master/data/deft\_files/dev/t1\_biology\_0\_0.deft}}
\end{quote}

Occasionally, terms may appear as verb phrases or other constituent phrases (e.g. adjective phrases):
\begin{quote}

Many crystals and solutions rotate the plane of polarization of light passing through them. Such substances are said to be \textit{optically active}.\footnote{\href{https://github.com/adobe-research/deft_corpus/blob/master/data/deft_files/dev/t3_physics_1_101.deft}{https://github.com/adobe-research/deft\_corpus/blob/master/data/deft\_files/dev/t3\_physics\_1\_101.deft}}

\end{quote}{}

Not including minor tokenization and sentence break fixes made during the shared task, there were no major modifications to the dataset before use in the DeftEval task. Complete details on the dataset is discussed in \newcite{spala:19}.

\section{DeftEval Task}\label{task_description}
The DeftEval shared task focused on fostering an expanded understanding of definition extraction for complex data, while keeping results comparable to traditional definition extraction tasks. The shared task, which ran from 4 Sept 2019 to 13 March 2020, was comprised of three subtasks: sentence classification, sequence labeling, and relation extraction. 

Participants were provided data in accordance with the DeftEval timeline:
\begin{itemize}
    \item \textbf{15 Aug 2019}: Practice data released
    \item \textbf{4 Sept 2019}: Training data released
    \item \textbf{19 Feb 2020}: Beginning of Subtask 1 and 2 evaluation period, unlabeled test data for Subtask 1 and 2 released
    \item \textbf{1 March 2020}: Beginning of Subtask 3 evaluation period, unlabeled test data for Subtask 3 released
    \item \textbf{13 March 2020}: End of competition period, labeled test data for all subtasks released
\end{itemize}

DeftEval was administered using \href{codalab.org}{Codalab Competitions}, a free, open-source platform for machine learning competitions and data challenges. Instructions for participation, including details about the annotation schema and information about the evaluation process, were made available during the practice and training periods of the shared task. All data and scripts were disseminated through the DEFT Github repository\footnote{\href{https://github.com/adobe-research/deft_corpus}{https://github.com/adobe-research/deft\_corpus}}, and linked to on the Codalab landing page\footnote{\href{https://competitions.codalab.org/competitions/22759}{https://competitions.codalab.org/competitions/22759}}. Inconsistencies in the data, bug fixes, and updates to tokenization errors were made regularly throughout the training period as participants surfaced questions and potential problems with the dataset.

\subsection{Subtask 1: Sentence classification}
In keeping with previous definition extraction tasks, we believed it to be important that the DeftEval shared task included a sentence classification subtask. Participants were provided a script\footnote{\href{https://github.com/adobe-research/deft_corpus/blob/master/task1_converter.py}{https://github.com/adobe-research/deft\_corpus/blob/master/task1\_converter.py}} to convert the existing dataset in its sequence-labeling format to individual sentences with a binary label indicating whether or not the sentence contained a definition. For the purposes of this subtask, only sentences which contained token sequences labeled as \textit{definition} in the annotated data were considered "definition sentences". Sentences containing only \textit{secondary definitions} were given a \textit{0} label. Unlabeled test data for the evaluation period was provided pre-converted in the sentence classification format; participants were only required to predict the label of each sentence at test time.

\subsection{Subtask 2: Sequence labeling}
Because of the nature of the expanded annotation in the DEFT corpus in comparison to previous definition extraction corpora, we believed it to be particularly important that the DeftEval shared task include a sequence labeling task. The DEFT corpus contains a large amount of auxiliary labels not previously seen in existing corpora, and the ability to reliably extract these labels could indicate a model with a broader cabability to identify complex and lengthy definition relationships. Ultimately, we hope for models that could both extract \textit{and} pair (as seen in Subtask 3\ref{subtask_3}) related term and definition labels together, but given the potential for these two tasks to be particularly difficult and the novel nature of the DEFT corpus, the tasks were split into two subtasks for DeftEval. 

The default format of the DEFT corpus data is similar to the CoNLL 2003 format \cite{conll:2003} and formatted as BIO data \cite{ramshaw:1995}. Therefore, participants did not need to convert any of the data for Subtask 2. At test time, participants were provided sentences where each line contained a token, source text file, lower char bound and upper char bound. Participants were required to predict only the appropriate BIO tag. 

\subsection{Subtask 3: Relation Extraction}\label{subtask_3}

The final DeftEval subtask focused on extracting individual term-definition pairs, as well as matching auxiliary labels to their appropriate root. Without having a precedent for this particular task in the definition extraction domain, it was difficult to predict the level of difficulty for this task. Given the nature of existing relation extraction tasks, we expected this task to be particularly complex.

Participants were again able use the default format of the DEFT data for training.  At test time, participants were provided sentences where each line contained a token, source text file, lower char bound, upper char bound, BIO tag, and tag ID. Participants were required to predict both the tag ID of the sister relation and the relationship tag. 

\subsection{Task Logistics}
All participants were required to use the provided data for training, but were allowed to add additional resources to their training data (e.g. other definition extraction resources, such as WCL, or general purpose models such as BERT). Additional hand-labeled definition extraction data, proprietary datasets, or unpublished data not universally accessible was explicitly disallowed. Participants were required to use the DEFT data as intended, using dev and training data \textit{only} for training and development. Though it was only made available during the evaluation periods, participants were also informed that test data was to be used solely for testing purposes.

During the training period, participants were allowed unlimited submissions to the Codalab practice and training phases. The practice phase evaluated against the sample "practice" data, consisting of three "dummy" files, one for each subtask. The training phase evaluated against the dev set data. Exact copies of these evaluation data were provided in the Github repository. Participants were able to submit as many subtasks together as desired, though the evaluation process required that \textit{all} files in a particular subtask be present for evaluation. In other words, participants must submit all files in the subtask evaluation set in order to be evaluated.

Evaluation scripts were also made available on the Github repository so that participants had access to understand and test the evalaution methods. Participants were encouraged to test their models locally on their own machines, then compare those results to those provided by the Codalab portal to ensure the formatting of their submissions were consistent with the Codalab requirements. Participants were also provided access to sample submissions via the Github repository for both the practice and training phases.

Each evaluation period ran for 12 days in which participants were able to submit their results through the Codalab submission process. Because of the nature of the relation extraction task potentially providing data for the sequence labeling task, the evaluation periods for Subtask 1 and 2 ran concurrently for 12 days, followed by the 12 day evaluation period for Subtask 3. Each participant was allowed 5 submissions per subtask, primarily to allow for any complications in the submission process. The vast majority of participants who submitted to the evaluation period used all 5 submissions. We suspect that many participants used these 5 submission attempts strategically in order to achieve higher results on the test sets, which should be taken into account for future tasks.

\section{Shared Task Results}

Over the course of the shared task competition period, 279 submissions were made through the Codalab training phase, with many participants additionally developing and testing their systems locally. The evaluation periods brought 217 submissions for Subtask 1, 220 for Subtask 2, and 113 for Subtask 3. Results are based on the evaluation metrics described in Section \ref{task_description}. 

\begin{figure*}
    \centering
    \includegraphics[width=\textwidth]{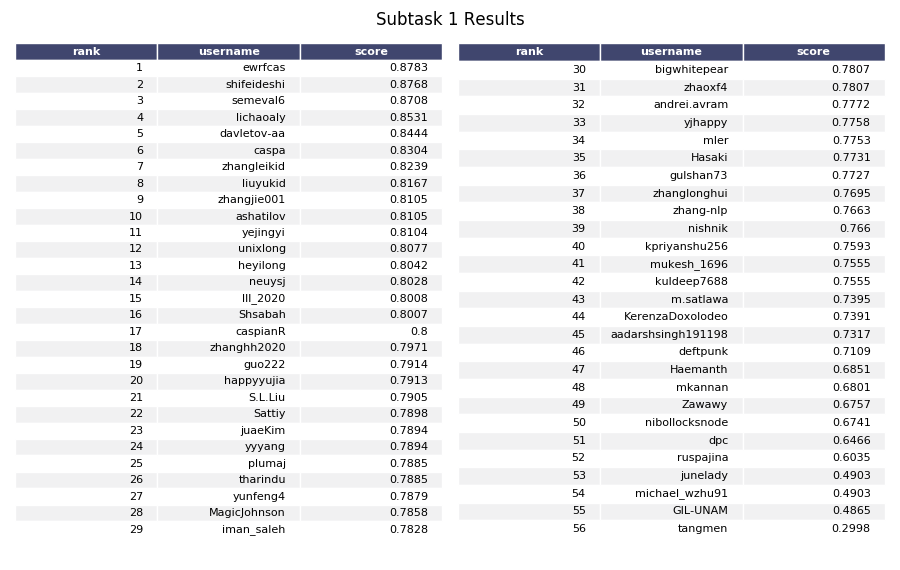}
    \caption{Subtask 1 results; 56 participants, 217 total successful submissions}
    \label{fig:subtask1_results}

    \centering
    \includegraphics[width=\textwidth]{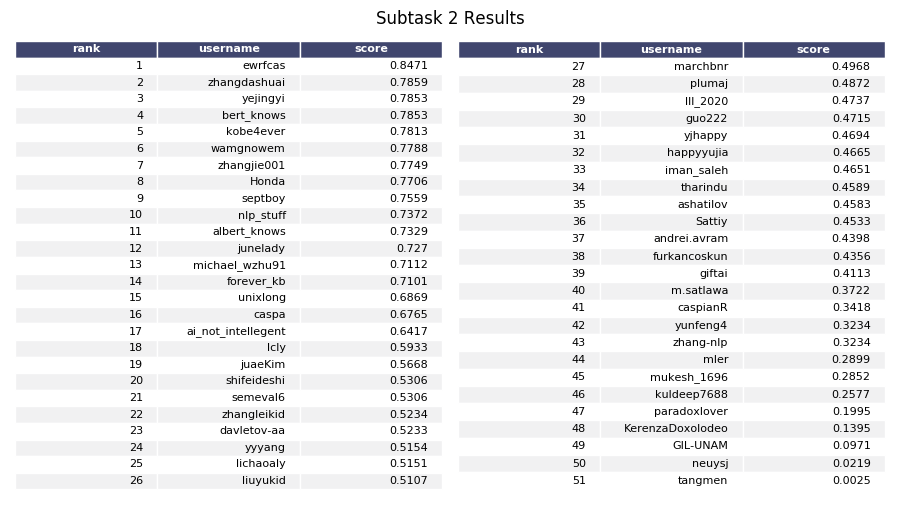}
    \caption{Subtask 2 results; 51 participants, 220 total successful submissions}
    \label{fig:subtask2_results }
    
    \centering
    \includegraphics[width=\textwidth]{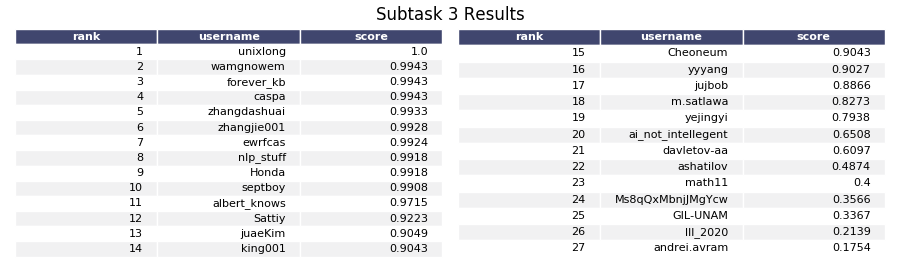}
    \caption{Subtask 3 results; 27 participants, 113 total successful submissions}
    \label{fig:subtask3_results}
    
\end{figure*}{}

\begin{table}[]
\begin{tabular}{p{.7cm}p{2.4cm}p{2.5cm}p{3.6cm}p{5cm}}
\toprule
\textbf{Rank} & \textbf{Team Name} & \textbf{Submission Name}           & \textbf{Reference} & \textbf{Models}                                                                           \\ \midrule
5             &  $^*$               & davletov-aa                        &   \newcite{davletov:2020}           & Multi-task BERT                                                                           \\
6 & ACNLP & caspa & \newcite{caspani:2020} & RoBERTa + Stochastic Weight Averaging
\\
12            & UNIXLONG           & unixlong                           & \newcite{xie:2020}  & Joint classification and sequence labeling pre-trained model with MLP and CRF layer       \\
16            & Cardiff University & Shsabah                            &   \newcite{jeawak:2020}           & BERT with two-step fine tuning                                                            \\
18            & BERTatDE           & zhangh2020                         &  \newcite{zhang:2020}            & BERT with BiLSTM + attention                                                              \\
26            & RGCL               & tharindu                           & \newcite{ranasinghe:2020}             & XLNet                                                                                     \\
32          & UPB & andrei.avram & \newcite{avram:2020} & RoBERTa with finetuning 
\\
40       & DSC IIT-ISM        & kpriyanshu256&   \newcite{singh:2020}           & BERT with fine-tuned language model                                                       \\
46            & DeftPunk           & deftpunk                           & \newcite{soboleva:2020}             & FastText and ELMo embeddings with RNN ensemble                                            \\
47            & TüKaPo             & haemanth                           &   \newcite{kannan:2020}           & Concatenated GloVe and on-the-fly POS embeddings with BiLSTM and 1D-Conv + MaxPool layers \\ \bottomrule
\end{tabular}
\caption{Subtask 1 submissions and system descriptions. \\ $^*$Team name not specified}
\label{subtask1_submissions}
\end{table}

\begin{table}[]
\begin{tabular}{@{}lllll@{}}
\toprule
\textbf{Rank} & \textbf{Team Name} & \textbf{Submission Name} & \textbf{Reference} & \textbf{Models}        \\ \midrule
23            &   $^*$                   & davletov-aa              &   \newcite{davletov:2020}           & BERT                   \\
27            & defx               & marchbnr                 & \newcite{hubner:2020} & CRF tagger             \\
34            & RGCL               & tharindu                 &  \newcite{ranasinghe:2020}            & XLNet - large          \\
37 & UPB & andrei.avram & \newcite{avram:2020} & RoBERTa + CRF with finetuning 
\\
\bottomrule
\end{tabular}
\caption{Subtask 2 submissions and system descriptions. \\ $^*$Team name not specified}
\label{subtask2_submissions}
\end{table}

\begin{table}[]
\begin{tabular}{@{}lllll@{}}
\toprule
\textbf{Rank} & \textbf{Team Name} & \textbf{Submission Name} & \textbf{Ref} & \textbf{Models}           \\ \midrule
1             & UNIXLONG           & unixlong                 & \newcite{xie:2020}  & BERT + hand crafted rules \\ 
4 & ACNLP & caspa & \newcite{caspani:2020} & Random Forest
\\
\bottomrule
\end{tabular}
\caption{Subtask 3 submissions and system descriptions.}
\label{subtask3_submissions}
\end{table}

Participants in DeftEval 2020 used a various array of methods for the three subtasks. For Subtasks 1 and 2 (see Tables \ref{subtask1_submissions} and \ref{subtask2_submissions}), many participants opted simply to use a pre-trained language model, especially BERT, RoBERTa, and XLNet (see \newcite{xie:2020}, \newcite{avram:2020}, \newcite{singh:2020}, \newcite{jeawak:2020}, \newcite{ranasinghe:2020}, and \newcite{davletov:2020}). Other implementations incorporated various LSTM layers in addition to the use of BERT or other transformer architecture (see \newcite{zhang:2020}). 

Participants were also particularly interested in multi-task solutions for Subtasks 1 and 2, to varying degrees of success (see \newcite{xie:2020}, \newcite{avram:2020}, \newcite{gordeev:2020}). \newcite{avram:2020} used pre-trained embeddings, then "projected the contextualized embeddings generated by the RoBERTa model in three vectors", making predictions for all three subtasks with a cohesive architecture.

Some ``classic" neural architectures were also used in participant submissions to Subtask 1, including \newcite{kannan:2020}'s LSTM, RNN, and ``hybrid" CNN solution, which leverages ``the implicit local feature-extraction performed by the convolutional layers to refine the final representations passed to the reccurant layer, which accounts for global features." \newcite{ranasinghe:2020} also used CNN and RNN methods on Subtask 1, in addition to other pretrained architectures.

Subtask 3 submissions (see Table \ref{subtask3_submissions}) were generally more simple in nature, perhaps due to the issues discussed in Section \ref{subtask3_difficulty}. \newcite{xie:2020} used a pre-trained BERT model with some rule-based layers, and \newcite{caspani:2020} used a random forest approach with additional post-processing.

In addition to an array of architectures, DeftEval also sparked conversation around the annotation schema. \newcite{jeawak:2020} point out that the complexity of the schema make this task, as expected, more difficult than previous definition extraction tasks. They also note that some annotations conflict with each other in the training and test sets, which may be due to the nature of the schema, as well as the complicated content of the documents annotated. We recognized this issue, and recommend that future definition extraction tasks and tasks using the DEFT data consider this. Others mentioned they found that their initial model explorations came up short; \newcite{singh:2020} mention that ``owing to the complexity of the DEFT corpus" their ``syntactically aware neural networks" performed lower than expected.

Overall, participants employed a variety of different models, especially those based on pre-trained transformer architectures for Subtasks 1 and 2 and more ``classic" approaches for Subtask 3. Participants also pointed out discrepancies and issues with the source dataset, which will be accounted for in future work.

\section{Complications}
\subsection{Ammendments to the DEFT Corpus}
The source data for DEFT was extracted by an automatic script that collected and sentence tokenized content from \href{cnx.org}{cnx.org}. While modern automatic tokenization methods generally work well in clean environments, content scraped from cnx.org often unintentionally includes raw urls, incorrectly parsed text, and errant html code. Because of this, the tokenization process for the textbook data was particularly error prone. Though our annotators and developers spent significant time manually adjusting tokenization errors, many still existed at the time of release on Github. Many participants discovered these tokenziation errors during training, which were then verified and fixed via Github issues. Because of the mixed nature of the source problem for tokenization errors, many caused incorrect sentence tokenization, which in turn meant that the actual number of sentences in the corpus was different from the reported number in \cite{spala:19}. With the existing tokenziation fixes at publication time, the textbook corpus contains 16056 sentences (695627 tokens) and 9439 2-3 sentence context windows.

\subsection{Subtask 3 difficulty} \label{subtask3_difficulty}
In organizing Subtask 3, we believed the task of matching term-definition pairs to be particularly difficult, especially considering the number of additional possible tags in the DEFT corpus. Given the precedence of previous performance on other more constrained relation extraction tasks (see also \newcite{lin:2016}, \newcite{shen:2016}, \newcite{sorokin:2017}, etc.), we predicted this task would see mostly low performing scores. 

However, participants in the subtask received a median F1 score of 0.9043, with 16 of the 27 participants scoring above 0.9. Upon further inspection, we realize this is due to the narrow search space of the three-sentence context windows. In designing the subtasks, we decided to keep the context windows in the test set so that the format of the data remained the same between released training data and test data and to minimize confusion for an already complicated task. Unfortunately, most context windows only include one set of term, definition, and matching auxiliary tags, meaning that this task was in fact particularly easy. F1 scores of 1 and near 1 make sense given the fact that models had to predict scenarios which in many cases only had one possible answer. Only cases that had more than one term-definition pair had a search space larger than 1. 

We recommend that future iterations of this particular task provide sentences in context with the entire document. Though this does not work with the current format of the DEFT corpus, providing annotated data in context with the rest of the source textbook is a trivial task. While context windows only show the range of three sentences and possible distance between term-definition pairs would remain within those three sentences, providing the full context of the textbook would mean that models would have to discern term-definition pairs that cross context windows or appear in neighboring sentences. Future tasks may also consider combining Subtasks 2 and 3 in order to create a more comprehensive and challenging task overall. 

\subsection{Concurrent subtask evaluation periods}
The evaluation periods for Subtasks 1 and 2 were scheduled to run concurrently in order to maximize the amount of time alloted for each subtask without extending the official SemEval evaluation period and to account for the data needs of Subtask 3. While Codalab's interface allowed for scheduling of two separate phases which had concurrent dates, the implementation in fact only allows for one phase. As a solution, we merged the two subtask evaluation phases into one phase which accepted both subtasks' submissions. Unfortunately, this also meant that individuals were now only provided 5 total submissions for \textit{both} subtasks. To handle this, participants were instructed to complete submissions for their Subtask 1 evaluation first, then contact the task organizers to open additional submissions for Subtask 2 evaluation, keeping their best scoring Subtask 1 results in their submission files so that both Subtask 1 and Subtask 2 scores would be available on the leaderboard. Participants were warned that if their submissions for Subtask 1 changed with their Subtask 2 data after their 5 alloted submissions, the results would be removed and they would be subject to removal from the leaderboard.

We recommend future tasks with multiple subtasks using Codalab to either run these subtasks on separate, non-overlapping dates, or to provide the total number of submissions needed for \textit{both} subtasks in the phase instructions, and remove any submissions that exceed the alloted submissions for the individual subtasks.

\section{Conclusion}
Overall, we believe the DeftEval shared task was a successful first exploration of definition extraction with the DEFT corpus. With more than 400 submissions and 149 registered participants, definition extraction as a complex, real-world task is clearly of interest to many. We believe that the results from this shared task provide a baseline for expectations for more complicated definition extraction tasks.

Though this task does not test a production-ready or real-life scenario, where models may have to predict term-definition pairs in a large document or in a wider context, the task does provide an initial look at how models perform on broad, non-standard definition extraction annotations. We recommend that future iterations of this task pay close attention to the effect of the DEFT corpus' three sentence context windows at test time, especially as it relates to the differences between an artificially narrow search space and documents ``in the wild". We also recommend that future research includes a critical examination of conflicting annotations in the DEFT corpus.

Overall, we believe DeftEval has surfaced questions and research directions relevant to expanding definition extraction into more complicated linguistic environments, where data is particularly messy and the natural language it stems from is particularly complex. DeftEval provided a launching point for the exploration of this type of problem in definition extraction, proved the field interest for such a task, and laid groundwork for future iterations of definition extraction tasks.

\bibliography{definition_extraction}
\bibliographystyle{coling}
\end{document}